\documentclass[11pt]{article}
\usepackage[preprint]{acl}
\usepackage[T1]{fontenc}
\usepackage{times}
\usepackage{latexsym}

\usepackage{url}
\usepackage{graphicx}
\usepackage{subcaption}
\usepackage{booktabs}
\usepackage{amsmath}
\usepackage{amssymb}
\usepackage{xcolor}
\usepackage{microtype}
\usepackage{placeins}
\title{When Opposite Premises Translate Identically: A Source-Conditioned Diagnostic for Multilingual Commonsense Items}

\author{Shuai Wang \quad Rui Hu \\
  Zhejiang University}
\date{}
\hypersetup{
  pdftitle={When Opposite Premises Translate Identically: A Source-Conditioned Diagnostic for Multilingual Commonsense Items},
  pdfauthor={Shuai Wang and Rui Hu},
  pdfsubject={A controlled diagnostic of polarity loss in translated commonsense premises},
  pdfkeywords={multilingual evaluation, machine translation, commonsense reasoning, translation quality estimation}
}

\begin{document}

\maketitle

\begin{abstract}
Multilingual reasoning comparisons depend on translated premises preserving the relations in their English sources. Automatic translation scores can rank source--target pairs, but an acceptance decision alone cannot certify a counterfactual benchmark field. We test a source-contrast diagnostic for ACCORD premises: reverse the English polarity, translate both arms independently, count identical target outputs, and score each arm against its own source. Across 71 matched question identities at two containing-item depths, five target languages, and three translation checkpoints, Swahili has 63, 113, and 67 exact collisions among 710 pairs per checkpoint. At a human-reference-calibrated threshold, round-trip NLI passes both arms of 3 of the 113 NLLB-3.3B collisions, while LaBSE passes 78. Each collision proves at least one unfaithful translation, but does not identify its arm. These controlled measurements do not establish screening accuracy on spontaneous translation errors or a language-by-hop reasoning effect; the study has no native equivalence labels or answer-model evaluation.

\end{abstract}

\begin{figure*}[t]
  \centering
  \includegraphics[width=\textwidth]{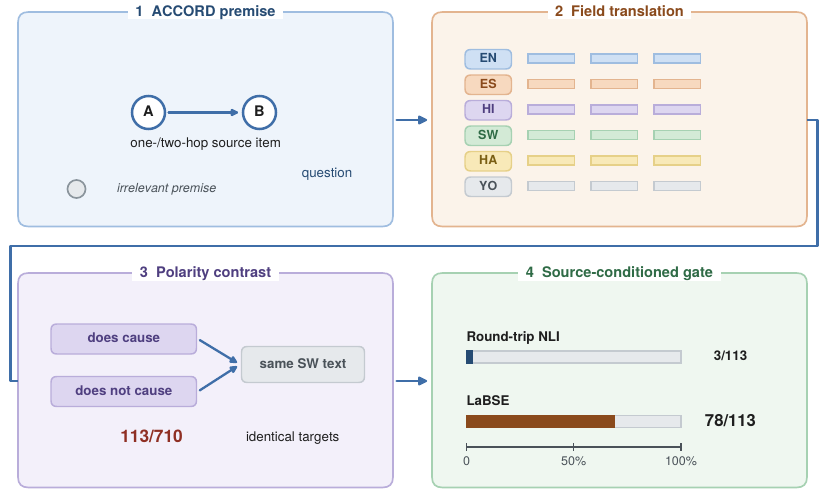}
  \caption{\textbf{Opposed ACCORD premises can collapse to the same target text.} The first panel locates a premise in a source item. The second shows field translation across the target languages. The third tests the polarity reversal and counts identical Swahili outputs. The fourth gives round-trip NLI and LaBSE decisions against each arm's generating source at their calibrated thresholds. A dual acceptance certifies at least one accepted unfaithful field without identifying its arm.}
  \label{fig:teaser}
\end{figure*}

\section{Introduction}
\label{sec:intro}

A multilingual comparison of reasoning depth requires the same counterfactual problem in every language. If a translated premise drops a negation, the target item supplies a different relation, and its answer may change before any reasoning model sees it. An apparent language gap can therefore combine comprehension, translation fidelity, and reasoning over the proposition graph. The present study examines the translation part of that dependency at the premise-field level.

ACCORD controls hops and distractors in English \citep{roewerdespres2024accord}. IrokoBench and XCOPA expose multilingual commonsense gaps on different item designs \citep{adelani2025irokobench,ponti2020xcopa}. Combining depth control and language coverage would require meaning-equivalent target items. Human judgments can establish equivalence, but scaling such review makes automatic screening attractive. Existing contrastive translation challenge sets, including ACES, show that metric scores can miss semantic changes \citep{amrhein2022aces}. What remains unknown for ACCORD construction is whether a field translator preserves the polarity of its short counterfactual premises, and how candidate screens behave on definite failures in those fields.

The diagnostic exploits a source contrast whose logical consequence is observable without treating an automatic metric as ground truth. We toggle the English polarity of an ACCORD premise and translate both arms independently. When their target strings are identical, one target cannot express both opposed sources. That collision proves at least one unfaithful translation, although it does not identify which source--target pair is wrong. The test is specific to a benchmark construction where premise polarity controls the relation available to the reasoner.

Figure~\ref{fig:teaser} summarizes the source-contrast diagnostic. We apply the contrast to matched ACCORD question identities in one- and two-hop containing items, using five premise fields per item. Two NLLB checkpoints and MADLAD-400 provide three translation conditions across Spanish, Hindi, Swahili, Hausa, and Yoruba. Four fixed scores then evaluate each collided target against its own generating English source. Their thresholds are set on separate FLORES human translations. This source-conditioned design avoids the automatic same-score result obtained by comparing both arms against one English source. The two item depths replicate the translation probe across source constructions; each premise field itself expresses one relation edge, so its collision rate is not a reasoning slope.

Round-trip entailment checks whether a back-translation preserves the English source, while direct multilingual entailment avoids a second translation step. LaBSE measures cross-lingual embedding similarity, and COMET-QE supplies a reference-free quality score \citep{conneau2018xnli,feng2022labse,rei2020wmt}. These instruments have different language coverage and different responses to short benchmark fields. We therefore report their decisions separately, at a stated human-reference retention rule, rather than combining them into a proposed fidelity certificate. The paired collision is the evidence that a source distinction was lost. Screen acceptance matters only when both generating-source comparisons pass, because then at least one accepted source--target pair must be unfaithful.

In Swahili, exact collisions occur with all three translators: 63, 113, and 67 among 710 opposed pairs per checkpoint. For NLLB-3.3B, round-trip NLI passes both source-conditioned arms in 3 of 113 collisions at its calibrated threshold, while LaBSE passes 78. A source-format comparison is confounded by batch-size differences: some inputs that were identical in both runs received different target outputs. The large Yoruba MADLAD collision count is dominated by repetitive output, a checkpoint-specific failure rather than evidence about Yoruba semantics.

The contribution is a reproducible, logically grounded collision certificate for translated ACCORD premise fields and a matched comparison of four source-conditioned screen decisions. Exact collisions reveal forward-translation failures without native labels; dual-accepted collisions place a lower bound on accepted unfaithful fields under the tested gates. We do not estimate translation-error detection precision or recall, validate complete multilingual items, or evaluate answer models. The language-by-hop reasoning hypothesis remains open until paired target items are judged for meaning equivalence.

\section{Related Work}
\label{sec:related}

\paragraph{Controlled and multilingual evaluation.}
ACCORD varies counterfactual reasoning hops and distractors in English \citep{roewerdespres2024accord}. XCOPA and IrokoBench provide multilingual causal and broader evaluation items \citep{ponti2020xcopa,adelani2025irokobench}. IrokoBench includes human-translated AfriXNLI and AfriMGSM items and evaluates translate-test alongside direct prompting \citep{adelani2025irokobench}. Global MMLU improves translated evaluation through professional and community review \citep{singh2025globalmmlu}. These designs show the value of multilingual item quality, but none supplies the same ACCORD proposition graph across controlled depths in our target languages. The present study probes polarity preservation in machine-translated ACCORD premises and the behavior of automatic screens on that contrast.

\paragraph{Translation quality and contrastive tests.}
Round-trip translation makes bilingual quality estimation an English comparison, with behavior depending on the reverse model and semantic metric \citep{moon2020rtt,zhuo2023rtt}. COMET-QE, CometKiwi, and xCOMET are reference-free alternatives, while LaBSE measures cross-lingual sentence similarity \citep{rei2020wmt,rei2022cometkiwi,guerreiro2024xcomet,feng2022labse}. ACES already uses contrastive challenge sets to expose metric failures that aggregate correlation hides \citep{amrhein2022aces}. Our controlled contrast concerns the forward translator as well as the screen: when opposed ACCORD premise sources yield the same target, at least one arm is unfaithful without requiring a metric or native judge to label it. Testing each arm against its generating source then gives a lower bound on accepted unfaithful fields. This certificate applies only to exact collisions; it does not assess the accuracy of a screen on other errors.

\paragraph{Inspectable language-model workflows.}
PaperPilot represents multi-turn scientific literature search as an executable graph of search operators and refines that graph using user feedback \citep{li2026multi}. Its emphasis on exposing intermediate operations is complementary to our diagnostic, which separates source editing, translation, collision detection, and source-conditioned scoring. The tasks differ: PaperPilot evaluates literature retrieval, whereas our study measures translation collisions and screen decisions.

\section{Method}
\label{sec:method}

\subsection{Paired ACCORD premise fields}
\label{sec:gate}
We select the 71 distinct ACCORD question identities with an eligible one-hop item and a matched two-hop item from the public generator snapshot \citep{roewerdespres2024accord}. Within each relation category, the eligible one-hop item with the earliest stable ID is selected per question; the matched two-hop item is selected by the same rule. Selection is fixed before translation or scoring. Each item has five premise fields. For the two-hop item, we take one relation edge from each premise field, rather than translating the whole proposition chain. Thus source depth describes the containing item; a field in either stratum expresses one relation. The same question identities occur at both depths.

For each premise $x$, a deterministic English edit toggles its first eligible \emph{does}, \emph{do}, or \emph{is} polarity marker. Denote the result by $\bar{x}$. The edit changes only that marker in the ACCORD template. We translate $x$ and $\bar{x}$ independently, without neighboring fields, using greedy decoding and a fixed output cap. The target languages are Spanish, Hindi, Swahili, Hausa, and Yoruba. We use NLLB-200 distilled 1.3B, NLLB-200 3.3B \citep{nllb2022}, and MADLAD-400 3B \citep{kudugunta2023madlad400}. The two NLLB checkpoints share a family; MADLAD supplies a separate architecture and training family. Field isolation makes the polarity contrast locatable but removes context that could resolve an ambiguous concept.

\subsection{A collision certificate and screen decisions}
\label{sec:collision-method}
Let $T_{m\ell}(x)$ be translator $m$'s output for source $x$ in language $\ell$. A collision occurs when the two target strings agree after case folding and whitespace normalization:
\begin{equation}
\begin{aligned}
c_{m\ell}(x)=\mathbf{1}\{&\operatorname{norm}(T_{m\ell}(x))\\
&=\operatorname{norm}(T_{m\ell}(\bar{x}))\}.
\end{aligned}
\label{eq:collision}
\end{equation}
Because $x$ and $\bar{x}$ express opposite source polarities, one common target cannot be meaning-equivalent to both. A collision therefore certifies at least one unfaithful source--target pair. It does not identify the faulty arm, and a noncollision does not certify either arm.

Each screen assigns a source-conditioned score $s(x,t)$ to English source $x$ and target $t$. For round-trip NLI, we back-translate $t$ with the corresponding translator and take the smaller bidirectional entailment probability between $x$ and the back-translation. Direct NLI takes the smaller bidirectional entailment probability between $x$ and $t$. Both use mDeBERTa-v3-base-mnli-xnli; its XNLI fine-tuning includes Spanish, Hindi, and Swahili but excludes Hausa and Yoruba \citep{conneau2018xnli}. LaBSE uses normalized cross-lingual embedding cosine \citep{feng2022labse}. WMT20 COMET-QE scores the source--target pair without a reference \citep{rei2020wmt}; its published encoder coverage excludes Yoruba. These are checkpoint-specific comparisons, not an evaluation of all current quality estimators.

For each language and screen, the first 100 aligned human FLORES development translations set the highest observed threshold $\tau$ that retains at least 90 references. Round-trip thresholds are refitted for each reverse translator; the other three screens use the same human-reference pairs across translator conditions. Thresholds are fixed before ACCORD scoring. We evaluate both collision arms against their \emph{own generating sources}. A dual acceptance is
\begin{equation}
\begin{aligned}
d_{m\ell}(x)={}&c_{m\ell}(x)\,\mathbf{1}\{s(x,T_{m\ell}(x))\geq\tau\}\,\\
&\mathbf{1}\{s(\bar{x},T_{m\ell}(\bar{x}))\geq\tau\}.
\end{aligned}
\label{eq:dual}
\end{equation}
Each dual-accepted collision certifies at least one accepted unfaithful field. This count is a lower bound on false acceptances among the tested fields, not a precision or recall estimate. We resample the 71 source question identities with their five premises together for uncertainty. Threshold resampling separately varies the 100 FLORES calibration references. Neither procedure supplies target-language equivalence labels.

\subsection{Secondary controls and complete items}
\label{sec:calibration-method}
Two distinct controls characterize score behavior away from exact collisions. The human-translated control pairs an English XNLI entailment hypothesis with its human target and with a separately authored contradiction hypothesis's human target from the same source group. It spans 200 aligned XNLI/AfriXNLI test groups per language \citep{conneau2018xnli,adelani2025irokobench}. The ACCORD choice control substitutes the translation of the least-similar sibling answer for an unverified original translation. Both are ranking controls, not labels for spontaneous translation errors. FLORES wrong-row and mechanical source-negation controls are reported in the appendix.

The original 24 one-hop items also support a complete-item sensitivity analysis: an item passes only if its five premises, question, and five choices all pass, and a multilingual set requires the same accepted English source ID in all five target languages. This conjunction tests screening yield, not item validity. No answer model is evaluated and no language-by-hop accuracy slope is estimated.

\section{Experiments}
\label{sec:experiments}

\subsection{Setup}
\label{sec:setup}
The main diagnostic contains 71 matched ACCORD question identities at each of two source-item depths. Five premise fields per item give 355 opposed source pairs per depth and target language. We translate both arms into Spanish, Hindi, Swahili, Hausa, and Yoruba with three checkpoints. An exact collision is independent of the four automatic screens. Screen decisions use each arm's own English source and thresholds fitted on separate FLORES human translations. No translated premise in the main diagnostic has a native equivalence label. Source-question bootstrap resampling keeps the five premises and two depths together; calibration-set resampling assesses threshold sensitivity.

\subsection{Which opposed sources produce the same target?}
\label{sec:polarity}
\begin{figure*}[t]
\centering
\includegraphics[width=\textwidth]{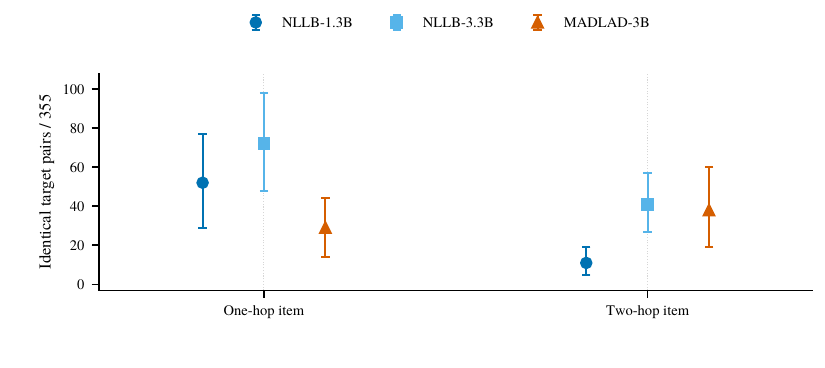}
\caption{\textbf{Exact Swahili target collisions.} Each point is the number of identical translations of opposed premises among 355 pairs at the indicated containing-item depth. Whiskers are percentile 95\% intervals from 2,000 source-question bootstrap draws, retaining all five premises. The two NLLB checkpoints share a model family; MADLAD-400 is a separate translator family.}
\label{fig:round7-collisions}
\end{figure*}

Table~\ref{tab:round7-collisions} reports exact target collisions. The pattern recurs in Swahili across translator families: NLLB-1.3B has 63 collisions, NLLB-3.3B has 113, and MADLAD-400 has 67 among 710 source pairs per translator. Because opposed source polarities cannot both match one target, each collision certifies at least one unfaithful field. The certificate does not determine whether the original or edited source was mistranslated. Figure~\ref{fig:round7-collisions} separates the one- and two-hop containing items; each premise field itself expresses one relation edge, so these counts are not a reasoning-depth effect.

One NLLB-1.3B Swahili example translates both \emph{refrigerator does not cause ham\_to\_cook} and \emph{refrigerator does cause ham\_to\_cook} as \emph{Tuseme kwamba friji haina kusababisha ham\_to\_cook}. The identical output cannot preserve both English propositions. It does not establish which of the two source--target pairs is faithful.

\begin{table*}[t]
\centering
\footnotesize
\caption{Exact target collisions and dual source-conditioned screen acceptance. Each depth and language has 355 opposed premise pairs from 71 questions. Collision counts are target-string identities. The four right columns count pairs for which both contradictory sources and their identical target pass the screen. RT thresholds are recalibrated for each reverse translator; other screens use the same FLORES human-reference threshold. All columns report counts, not translation-error rates.}
\label{tab:round7-collisions}
\begin{tabular}{llrrrrrr}
\toprule
Translator & Language & 1-hop & 2-hop & RT & Direct & LaBSE & QE \\
\midrule
NLLB-1.3B & Spanish & 1 & 0 & 0 & 0 & 0 & 0 \\
NLLB-1.3B & Hindi & 1 & 0 & 0 & 0 & 0 & 0 \\
NLLB-1.3B & Swahili & 52 & 11 & 4 & 0 & 43 & 11 \\
NLLB-1.3B & Hausa & 7 & 0 & 3 & 7 & 3 & 5 \\
NLLB-1.3B & Yoruba & 2 & 0 & 2 & 2 & 1 & 2 \\
\addlinespace
NLLB-3.3B & Spanish & 2 & 3 & 1 & 0 & 0 & 1 \\
NLLB-3.3B & Hindi & 0 & 3 & 0 & 0 & 2 & 0 \\
NLLB-3.3B & Swahili & 72 & 41 & 3 & 0 & 78 & 28 \\
NLLB-3.3B & Hausa & 3 & 3 & 1 & 4 & 1 & 4 \\
NLLB-3.3B & Yoruba & 3 & 0 & 2 & 3 & 3 & 3 \\
\addlinespace
MADLAD-3B & Spanish & 0 & 0 & 0 & 0 & 0 & 0 \\
MADLAD-3B & Hindi & 5 & 9 & 2 & 0 & 5 & 5 \\
MADLAD-3B & Swahili & 29 & 38 & 14 & 0 & 24 & 7 \\
MADLAD-3B & Hausa & 9 & 22 & 27 & 25 & 3 & 8 \\
MADLAD-3B & Yoruba & 198 & 111 & 309 & 309 & 1 & 144 \\
\bottomrule
\end{tabular}
\end{table*}

The cross-family replication has a limit. MADLAD-400's Yoruba output is often degenerate: one repeated target string accounts for 404 of its 1,420 Yoruba arm outputs. Its 309 Yoruba collisions consequently describe this checkpoint and decoding configuration, rather than a subtle omission of a polarity marker. The Hausa mode appears 49 times. We retain these outputs as observed failures and avoid treating the corresponding counts as representative language error rates.

\subsection{Do source-conditioned screens pass both arms?}
\label{sec:screen-results}
\begin{figure*}[t]
\centering
\includegraphics[width=\textwidth]{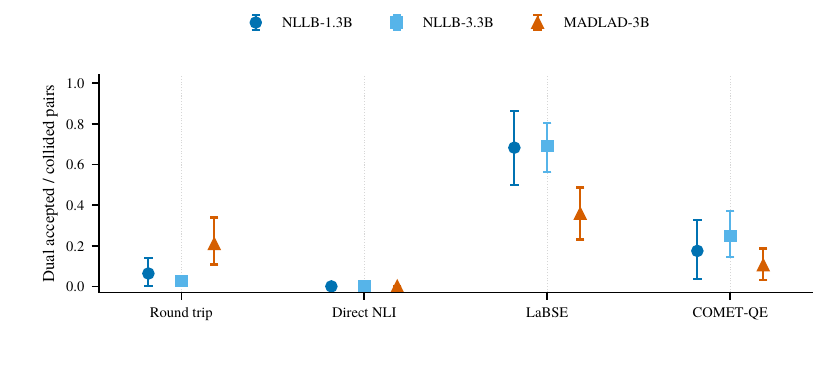}
\caption{\textbf{Swahili dual acceptance among exact collisions.} Both translated arms must pass against their own English sources. Points are observed fractions; whiskers are percentile 95\% intervals from 2,000 source-question bootstrap draws at frozen thresholds. Calibration uncertainty is analyzed separately. A dual-accepted collision proves that at least one accepted pair is unfaithful, without identifying which arm. The chart does not estimate screen precision or recall on naturally occurring errors.}
\label{fig:round7-dual}
\end{figure*}

Table~\ref{tab:round7-collisions} also gives dual acceptances, where each arm is scored against the English source that generated it. Comparing both identical targets with the same source would give a deterministic screen the same score twice, irrespective of fidelity. With source-conditioned comparisons, Swahili round-trip NLI dual-accepts 4 of 63 NLLB-1.3B collisions and 3 of 113 NLLB-3.3B collisions. LaBSE dual-accepts 43 and 78. For MADLAD-400 Swahili, the corresponding counts are 14 of 67 for round-trip NLI and 24 for LaBSE. In the degenerate MADLAD Yoruba condition, round-trip NLI passes both arms of all 309 collisions at a near-zero threshold. Figure~\ref{fig:round7-dual} compares the Swahili decisions at the fixed reference-calibrated operating points.

The FLORES thresholds are estimated from 100 references per language and screen. Bootstrap refits move some dual-acceptance counts, while the exact collision counts do not depend on threshold choice. Section~\ref{app:uncertainty} describes the uncertainty construction and reports selected numerical intervals. Direct NLI has no Swahili dual acceptances in these conditions, but its retention of faithful ACCORD fields is unknown. The comparison characterizes these checkpoints and this field-isolated translation protocol. It does not rank practical screening systems on native-judged ACCORD translations.

\subsection{Can the source-format comparison isolate an effect?}
\label{sec:normalization}
\begin{figure*}[t]
\centering
\includegraphics[width=\textwidth]{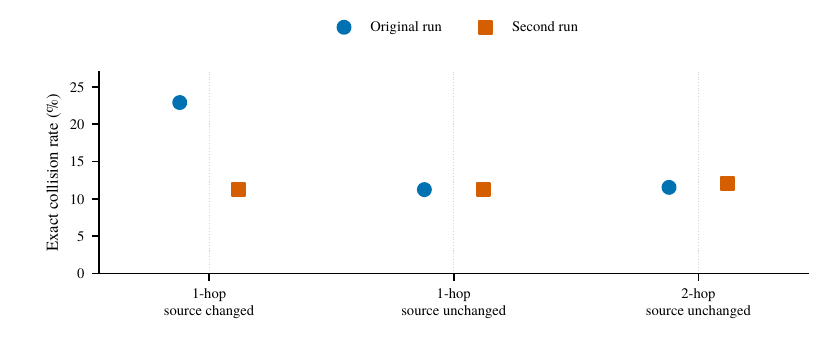}
\caption{\textbf{Swahili source-format comparison with unchanged-input controls.} Points give exact collision rates in two runs of the same NLLB-3.3B checkpoint. Only the first group has modified English sources; batch size also changed between runs. The unchanged two-hop group shows output variability under that execution change.}
\label{fig:input-control}
\end{figure*}

Underscore-delimited relation names appear in 275 of 355 one-hop premise pairs and in none of the 355 two-hop pairs. The depth strata therefore differ in source format. We compared the Swahili NLLB-3.3B run with a second run in which underscores were replaced by spaces. The original run used batches of 32 and the second used batches of 16, with the same checkpoint revision and greedy decoding. Figure~\ref{fig:input-control} separates edited from unchanged sources. Among the 275 edited one-hop pairs, collisions changed from 63 to 31. Among the 355 unchanged two-hop pairs, collisions changed from 41 to 43; three of their 710 unchanged source arms received different target strings. One of 160 unchanged one-hop arms also changed output, although its pair's collision status did not change. Thus the two-hop discrepancy is an execution difference, not an effect of replacing underscores. The runs do not isolate the effect of source formatting on the edited one-hop pairs. The cross-checkpoint collision certificate in Table~\ref{tab:round7-collisions} does not rely on this comparison.

\subsection{What do the remaining controls establish?}
\label{sec:secondary}
The human-translated XNLI/AfriXNLI control compares independently authored entailment and contradiction hypotheses. Its targets are human translations, so it does not exercise forward MT errors on ACCORD. The sibling-choice control uses unverified MT positives and deliberately dissimilar alternatives. Their ranking and calibration results, plus the 24-item all-field conjunction, appear in Section~\ref{app:controls} and Section~\ref{app:gate-sensitivity}. They cannot supply missing equivalence labels or an accepted multilingual benchmark. The available generation timing excludes screening and native review; cost per validated item is undefined because none was validated.

\section{Secondary controls and operating points}
\label{app:controls}
\begin{figure*}[t]
\centering
\includegraphics[width=\textwidth]{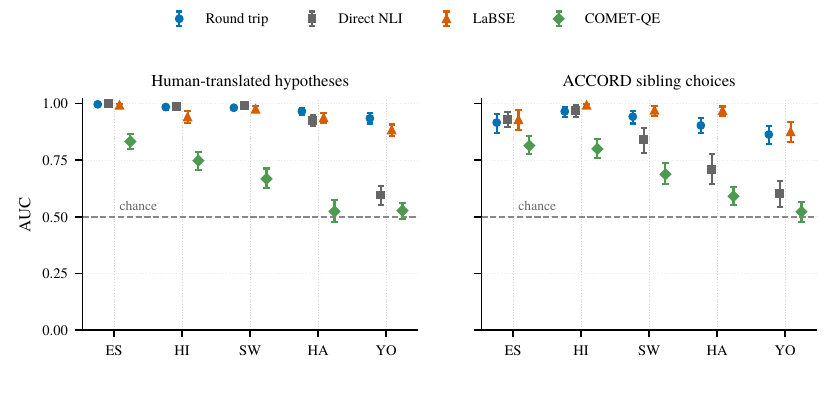}
\caption{\textbf{Ranking on secondary controls.} Left: human-translated entailment versus independently authored contradiction hypotheses. Right: original versus least-similar sibling-target ACCORD choices. Points are AUC; whiskers resample source groups or ACCORD items. The grey line is chance ranking. Neither control labels natural translation errors.}
\label{fig:paired-auc}
\end{figure*}

Table~\ref{tab:human-controls} compares the four screens on 200 aligned XNLI/AfriXNLI test groups per language. Each negative target is the human translation of a separately authored contradiction hypothesis from the same English premise group. These pairs test ranking on distinct sentences, rather than detection of forward-translation errors. Their English hypotheses have median character similarity 0.49; the XNLI-tuned NLI checkpoint may be advantaged by the task format. Figure~\ref{fig:paired-auc} also reports the original versus least-similar sibling-choice control. Original machine translations in this control are not native-verified.
\begin{table}[t]
\centering
\footnotesize
\caption{AUC for an English entailment hypothesis paired with its human translation versus a human translation of a contradiction-labeled hypothesis from the same premise. The 200 source groups are aligned across the five target languages. Higher is better.}
\label{tab:human-controls}
\begin{tabular}{lrrrrr}
\toprule
Screen & ES & HI & SW & HA & YO \\
\midrule
Round trip & 1.00 & 0.98 & 0.98 & 0.97 & 0.93 \\
Direct NLI & 1.00 & 0.99 & 0.99 & 0.93 & 0.60 \\
LaBSE & 0.99 & 0.94 & 0.98 & 0.94 & 0.88 \\
COMET-QE & 0.83 & 0.75 & 0.67 & 0.52 & 0.53 \\
\bottomrule
\end{tabular}
\end{table}

Table~\ref{tab:afrixnli-cv} contrasts FLORES-calibrated and five-fold XNLI/AfriXNLI-calibrated decisions for the human control. The original FLORES reference, wrong-row, and mechanical-negation results are in Table~\ref{tab:auc} and Table~\ref{tab:calibrated}. The latter edit inserts ``not'' after an English copula; inspection reveals double negation and scope changes, including ``not non-diabetic'' and ``not doing nothing.'' It is a fixed-edit sensitivity probe rather than an adjudicated translation-error set.
\begin{table*}[t]
\centering
\footnotesize
\caption{Aligned XNLI/AfriXNLI decisions under the FLORES development threshold and five-fold in-domain calibration. Each positive and altered-target column has 200 human-translated pairs. Each fold's in-domain threshold uses only the other four folds' positive targets. Counts are accepted pairs, not ACCORD validity labels.}
\label{tab:afrixnli-cv}
\begin{tabular}{llrrrr}
\toprule
& & \multicolumn{2}{c}{FLORES threshold} & \multicolumn{2}{c}{In-domain folds} \\
Language & Screen & Positive & Altered & Positive & Altered \\
\midrule
Spanish & Round trip & 188 & 1 & 178 & 0 \\
Spanish & Direct NLI & 190 & 0 & 177 & 0 \\
Spanish & LaBSE & 180 & 2 & 179 & 1 \\
Spanish & COMET-QE & 176 & 111 & 179 & 115 \\
\addlinespace
Hindi & Round trip & 181 & 5 & 180 & 3 \\
Hindi & Direct NLI & 176 & 2 & 178 & 5 \\
Hindi & LaBSE & 142 & 1 & 178 & 19 \\
Hindi & COMET-QE & 160 & 99 & 178 & 140 \\
\addlinespace
Swahili & Round trip & 179 & 7 & 179 & 10 \\
Swahili & Direct NLI & 183 & 4 & 181 & 4 \\
Swahili & LaBSE & 139 & 3 & 179 & 13 \\
Swahili & COMET-QE & 149 & 109 & 180 & 157 \\
\addlinespace
Hausa & Round trip & 178 & 18 & 176 & 19 \\
Hausa & Direct NLI & 183 & 58 & 180 & 42 \\
Hausa & LaBSE & 90 & 3 & 180 & 35 \\
Hausa & COMET-QE & 155 & 168 & 180 & 184 \\
\addlinespace
Yoruba & Round trip & 185 & 55 & 178 & 39 \\
Yoruba & Direct NLI & 191 & 180 & 180 & 164 \\
Yoruba & LaBSE & 156 & 32 & 181 & 65 \\
Yoruba & COMET-QE & 195 & 200 & 179 & 168 \\
\addlinespace
\bottomrule
\end{tabular}
\end{table*}

\begin{table*}[t]
\centering
\footnotesize
\caption{Round-trip NLI thresholds fitted separately to each reverse translator. Each threshold is the highest observed score retaining at least 90 of 100 aligned FLORES development human translations. Values are measured point estimates; target-language field retention is not known.}
\label{tab:rt-thresholds}
\begin{tabular}{lrrrrr}
\toprule
Reverse translator & Spanish & Hindi & Swahili & Hausa & Yoruba \\
\midrule
NLLB-1.3B & 0.162803 & 0.031874 & 0.025652 & 0.006631 & 0.001748 \\
NLLB-3.3B & 0.135743 & 0.023931 & 0.028563 & 0.007938 & 0.001714 \\
MADLAD-3B & 0.446000 & 0.023056 & 0.007618 & 0.004268 & 0.000551 \\
\bottomrule
\end{tabular}
\end{table*}

\begin{table*}[!h]
\centering
\footnotesize
\caption{Held-out AUC for distinguishing human references from wrong-row targets (M) and source-negated pairs (N). Higher is better; the negation comparison uses the corresponding 39 unaltered references.}
\label{tab:auc}
\begin{tabular}{lrrrrrrrr}
\toprule
& \multicolumn{2}{c}{Round trip} & \multicolumn{2}{c}{Direct NLI} & \multicolumn{2}{c}{LaBSE} & \multicolumn{2}{c}{COMET-QE} \\
Language & M & N & M & N & M & N & M & N \\
\midrule
Spanish & 0.98 & 0.93 & 1.00 & 0.94 & 1.00 & 0.68 & 0.96 & 0.65 \\
Hindi & 0.99 & 0.96 & 1.00 & 0.89 & 1.00 & 0.65 & 0.94 & 0.55 \\
Swahili & 0.99 & 0.90 & 1.00 & 0.87 & 1.00 & 0.65 & 0.84 & 0.54 \\
Hausa & 0.97 & 0.89 & 0.98 & 0.78 & 1.00 & 0.60 & 0.75 & 0.54 \\
Yoruba & 0.86 & 0.68 & 0.74 & 0.48 & 1.00 & 0.53 & 0.54 & 0.54 \\
\bottomrule
\end{tabular}
\end{table*}

For the mechanical negation pairs, replacing each back-translation with the unaltered English source yields NLI AUC 1.00. This identity-positive upper bound removes reverse translation and is easier than screening. COMET-QE median scores for held-out references, wrong-row targets, and source negations are -0.296, -0.665, and -0.351 in Hausa, and -1.606, -1.638, and -1.638 in Yoruba. These describe the tested WMT20 checkpoint only.
On the same Yoruba control, a conjunction of the calibrated round-trip NLI and LaBSE gates retains 75 of 100 held-out human references while accepting 22 of 39 mechanically negated source pairs. The control's edit defects prevent treating those decisions as translation-error precision.
\section{Complete-item and field-type sensitivity}
\label{app:gate-sensitivity}
The 24 one-hop items were translated field by field with NLLB-1.3B. A rendering passes only when all five premises, the question, and five choices pass. The FLORES-calibrated round-trip gate accepts 69 of 120 renderings; G27\_1\_A alone passes in all five languages, although its Yoruba question back-translates without the queried referent. This is a decision count, not validated yield. A 0.05 threshold floor removes that set and retains only 48 of 100 held-out Yoruba human references. The accepted set has not been validated by target-language judgments.

Table~\ref{tab:field-counts} gives field-type decisions. Spanish direct NLI retains 89 of 100 held-out FLORES references but passes 33 of 120 ACCORD choices, 107 premises, and 22 questions. The least-similar Yoruba sibling passes round-trip NLI in 114 of 120 cases, versus 119 original targets. These passes do not verify meaning.
\begin{table*}[t]
\centering
\footnotesize
\caption{Calibrated ACCORD field acceptance by type. Each language--screen pair has 120 premise fields, 24 questions, and 120 choices; counts are screen decisions without target-language equivalence labels.}
\label{tab:field-counts}
\begin{tabular}{llrrr}
\toprule
Language & Screen & Premise & Question & Choice \\
\midrule
Spanish & Round trip & 118 & 23 & 102 \\
Spanish & Direct NLI & 107 & 22 & 33 \\
Spanish & LaBSE & 64 & 23 & 73 \\
Spanish & COMET-QE & 56 & 10 & 67 \\
\addlinespace
Hindi & Round trip & 114 & 23 & 116 \\
Hindi & Direct NLI & 118 & 23 & 117 \\
Hindi & LaBSE & 88 & 19 & 102 \\
Hindi & COMET-QE & 53 & 15 & 61 \\
\addlinespace
Swahili & Round trip & 114 & 21 & 115 \\
Swahili & Direct NLI & 104 & 24 & 87 \\
Swahili & LaBSE & 111 & 15 & 86 \\
Swahili & COMET-QE & 42 & 15 & 68 \\
\addlinespace
Hausa & Round trip & 111 & 19 & 114 \\
Hausa & Direct NLI & 119 & 23 & 119 \\
Hausa & LaBSE & 86 & 18 & 78 \\
Hausa & COMET-QE & 111 & 18 & 110 \\
\addlinespace
Yoruba & Round trip & 117 & 22 & 119 \\
Yoruba & Direct NLI & 120 & 22 & 120 \\
Yoruba & LaBSE & 88 & 20 & 72 \\
Yoruba & COMET-QE & 119 & 24 & 120 \\
\addlinespace
\bottomrule
\end{tabular}
\end{table*}

At a common round-trip threshold of 0.80, 148 premises, 44 questions, and 222 choices fail among 120 renderings. Omitting choices admits 38 renderings and no complete set; premises alone admit 47 and one set, with questions and choices unchecked. NLLB-3.3B admits five full renderings versus four for 1.3B. These uncalibrated decisions do not certify meaning.
\section{English-side discrepancy audit}
\label{app:audit}
For each original-gate status, one author reviewed ten premises, ten questions, and ten choices in SHA-256 order of source ID, language, and field name, comparing English sources with back-translations. Apparent meaning change, plausible paraphrase, and ambiguous labels numbered 25, four, and one among rejected fields and eight, 20, and two among accepted fields. Gate-based selection prevents population precision or recall estimates. No target-language text was judged, and this audit does not cover calibrated thresholds.
\section{Resampling and uncertainty}
\label{app:uncertainty}
For the main diagnostic, 2,000 bootstrap draws sample the 71 ACCORD question identities with replacement and retain their five premises at both depths. We report percentile 95\% intervals using the 2.5th and 97.5th percentiles of those draws. Collision intervals derive from clustered counts; dual-acceptance fraction intervals condition on draws containing a collision. A separate 2,000-draw bootstrap samples the 100 FLORES development scores and recomputes the highest observed threshold retaining at least 90 human references. This propagates calibration uncertainty into the dual-acceptance count. The plotted intervals summarize source-question resampling at the frozen thresholds; the selected numerical intervals below separately describe threshold sensitivity.

For Swahili NLLB-3.3B, source-question resampling gives collision-count intervals of [48, 98] in one-hop containing items and [27, 57] in two-hop items. Refitting the FLORES threshold gives a one-hop dual-acceptance count interval of [1, 13] for round-trip NLI and [23, 69] for LaBSE; their fixed-threshold point counts are 3 and 53. These calibration intervals do not quantify fidelity among noncolliding targets.

The source-format bootstrap conditions on runs with different batch sizes, so it does not identify a normalization effect (Section~\ref{sec:normalization}).

Across 2,000 FLORES-threshold resamples, the complete six-language set count is zero, one, or at least two in 421, 1,127, and 452 draws; the identified set survives 74.3\%. These are decision-stability counts, not validation probabilities.

\FloatBarrier

\section{Conclusion}
\label{sec:conclusion}
Opposed ACCORD premise sources sometimes receive identical translations in two NLLB checkpoints and MADLAD-400. This provides a source-contrast certificate of lost polarity in at least one arm of each collision. Under source-conditioned scoring, the tested round-trip NLI gate rejects at least one arm of most Swahili NLLB collisions, whereas LaBSE passes both arms more often at its reference-calibrated threshold. The result is a controlled diagnosis of field translation and these screen checkpoints. It does not establish translation-error detection accuracy on naturally occurring errors, produce a validated multilingual benchmark, or test whether language changes the accuracy slope with reasoning depth.

\section*{Limitations}
The collision diagnostic pairs edited ACCORD premise fields from 71 question identities at two containing-item depths. A field in either stratum expresses one relation edge. We do not translate complete items in context or measure answer-model accuracy. The English polarity edit is a controlled intervention, not a sample of spontaneous MT errors. Exact target identity proves that one arm is unfaithful; it cannot identify that arm, quantify errors among noncollisions, or certify any original benchmark item. No target-language speaker has judged the translated premises, so screen precision, recall, and valid item yield remain unknown.

The two NLLB checkpoints share a translation family. MADLAD-400 provides a separate family but emits repetitive Yoruba outputs under the tested decoding configuration; its large Yoruba collision count should not be generalized to Yoruba translation. The source-format comparison changed batch size as well as text and produced different outputs for unchanged inputs. It cannot estimate a normalization effect. No contextual translator or LLM translator was evaluated. Translation runtime was not measured end to end with screening and human review, and cost per valid item is undefined.

The four screens are fixed public checkpoints rather than a survey of current quality estimation. The mDeBERTa NLI model lacks Hausa and Yoruba XNLI fine-tuning, and the WMT20 COMET-QE encoder's published coverage excludes Yoruba. We did not evaluate CometKiwi, xCOMET, an African-language NLI checkpoint, or an LLM semantic judge. Round-trip NLI mixes reverse-translation and inference behavior. Thresholds come from 100 FLORES human sentences per language, and decision counts can move when that calibration sample is resampled. The secondary XNLI/AfriXNLI hypotheses are independently authored and lexically distinct, while the choice probe deliberately selects a dissimilar sibling and uses unverified MT positives. Neither control substitutes for native judgments of ACCORD translation errors.
The sentence and ACCORD field controls also use different source templates and perturbations, so their outcomes do not isolate a field-structure effect or establish an improvement over ACES-style contrastive evaluation.

\bibliography{main}

\appendix
\section{Detailed operating-point tables}
\begin{table*}[!h]
\centering
\footnotesize
\caption{Held-out screen decisions after per-language thresholds are fixed to admit at least 90 of 100 FLORES dev human references. Columns count accepted ordinary devtest references, disjoint short devtest references, deliberately mismatched translations, source-negated references, and full ACCORD renderings. Their denominators are 100, 100, 100, 39, and 24 per language and screen. Threshold scales differ by screen; counts are decisions, not ACCORD meaning labels.}
\label{tab:calibrated}
\begin{tabular}{llrrrrrr}
\toprule
Language & Screen & $\tau$ & Devtest & Short & Mismatch & Negated & ACCORD \\
\midrule
Spanish & Round trip & 0.1628 & 91 & 94 & 0 & 12 & 10 \\
Spanish & Direct NLI & 0.9887 & 89 & 81 & 0 & 5 & 0 \\
Spanish & LaBSE & 0.8412 & 82 & 70 & 0 & 26 & 0 \\
Spanish & COMET-QE & -0.0456 & 93 & 96 & 18 & 38 & 0 \\
\addlinespace
Hindi & Round trip & 0.0319 & 96 & 87 & 0 & 14 & 16 \\
Hindi & Direct NLI & 0.6415 & 94 & 90 & 0 & 15 & 19 \\
Hindi & LaBSE & 0.8038 & 89 & 81 & 0 & 29 & 3 \\
Hindi & COMET-QE & 0.0239 & 89 & 95 & 18 & 33 & 0 \\
\addlinespace
Swahili & Round trip & 0.0257 & 91 & 84 & 0 & 13 & 14 \\
Swahili & Direct NLI & 0.4862 & 95 & 83 & 0 & 13 & 1 \\
Swahili & LaBSE & 0.8224 & 96 & 77 & 0 & 28 & 2 \\
Swahili & COMET-QE & -0.4061 & 83 & 80 & 37 & 31 & 1 \\
\addlinespace
Hausa & Round trip & 0.0066 & 91 & 85 & 4 & 20 & 10 \\
Hausa & Direct NLI & 0.0041 & 96 & 88 & 8 & 26 & 22 \\
Hausa & LaBSE & 0.7893 & 85 & 73 & 0 & 27 & 1 \\
Hausa & COMET-QE & -0.9226 & 92 & 86 & 77 & 33 & 10 \\
\addlinespace
Yoruba & Round trip & 0.0017 & 86 & 84 & 43 & 30 & 19 \\
Yoruba & Direct NLI & 0.0007 & 92 & 90 & 88 & 37 & 22 \\
Yoruba & LaBSE & 0.5882 & 85 & 75 & 0 & 28 & 2 \\
Yoruba & COMET-QE & -1.8187 & 90 & 100 & 92 & 35 & 23 \\
\bottomrule
\end{tabular}
\end{table*}

\section{Reproducibility}
\label{app:repro}
Eligible ACCORD one-hop records have a nonnull pairing, no negative-pairing flip, and one statement per A--E group. We select the earliest stable ID per question and relation category, then the earliest matching two-hop ID and one pairing-selected edge per field, fixing 71 identities. The English edit toggles the first eligible marker among \emph{does not}, \emph{does}, \emph{do not}, \emph{do}, \emph{is not}, and \emph{is}. All 710 pairs differ only there, although two source concepts contain a later \emph{do}. Identity uses case folding and whitespace collapse; source normalization replaces underscores with spaces. Each target is scored against its generating source.

The first 100 aligned FLORES development references set thresholds; reverse translation is refitted per checkpoint (Tables~\ref{tab:rt-thresholds} and~\ref{tab:calibrated}). XNLI/AfriXNLI folds use source groups. Forward translation uses half precision, greedy decoding, and a 96-token cap. The source-selection identifiers are supplied as an ancillary CSV file. The frozen checkpoint revisions are listed below. Full per-arm translation and score logs are not included in this preprint's ancillary files.

\paragraph{Checkpoint revisions.}
The public checkpoints and their frozen Hugging Face revisions are:
\begin{itemize}
\item NLLB-200 distilled 1.3B:\\
\texttt{7be3e24664b38ce1cac2}\allowbreak\texttt{9b8aeed6911aa0cf0576}.
\item NLLB-200 3.3B:\\
\texttt{1a07f7d195896b2114af}\allowbreak\texttt{cb79b7b57ab512e7b43e}.
\item MADLAD-400 3B:\\
\texttt{fa184c675da0b5c9e1c8}\allowbreak\texttt{694fccd4e12e2d422094}.
\item mDeBERTa-v3-base-mnli-xnli:\\
\texttt{8adb042d524ecd5c26d3}\allowbreak\texttt{e3ba0e3fbcf7e2d0864c}.
\item LaBSE:\\
\texttt{836121a0533e5664b21c}\allowbreak\texttt{7aacc5d22951f2b8b25b}.
\item WMT20 COMET-QE:\\
\texttt{2e7ffc84fb67d99cf925}\allowbreak\texttt{06611766463bb9230cfb}.
\end{itemize}

\section{Computational resources}
The translation and scoring experiments used H100 accelerators. Runtime was not recorded consistently across all components, so a reliable total GPU-hour estimate is unavailable.

\end{document}